\documentclass[runningheads]{llncs}
\usepackage{graphicx}
\usepackage{amsmath,amssymb} 
\usepackage{color}

\usepackage{algorithm}
\usepackage{algpseudocode}
\usepackage{xcolor}
\usepackage{makecell}
\usepackage{float}

\newcommand{\eg}{\textit{e}.\textit{g}.}

\usepackage{nicefrac}       
\usepackage{microtype}      
\usepackage{multirow}
\usepackage{verbatim}
\usepackage{color}
\usepackage{float}
\usepackage{enumitem}
\usepackage{booktabs}
\usepackage{tabulary,multirow,overpic,xcolor}
\usepackage[caption=false]{subfig}

\usepackage[british,UKenglish,USenglish,english,american]{babel}
\usepackage[pagebackref=true,breaklinks=true,letterpaper=true,colorlinks,bookmarks=false]{hyperref}

\newcommand{\ve}[1]{\mathbf{#1}} 
\newcommand{\ma}[1]{\mathrm{#1}} 

\newcolumntype{x}[1]{>{\centering\arraybackslash}p{#1pt}}

\newcommand{\app}{\raise.17ex\hbox{$\scriptstyle\sim$}}

\newcolumntype{x}[1]{>{\centering\arraybackslash}p{#1pt}}

\newlength\savewidth\newcommand\shline{\noalign{\global\savewidth\arrayrulewidth
  \global\arrayrulewidth 1pt}\hline\noalign{\global\arrayrulewidth\savewidth}}
\newcommand{\tablestyle}[2]{\setlength{\tabcolsep}{#1}\renewcommand{\arraystretch}{#2}\centering\footnotesize}

\makeatletter\renewcommand\paragraph{\@startsection{paragraph}{4}{\z@}
  {.5em \@plus1ex \@minus.2ex}{-.5em}{\normalfont\normalsize\bfseries}}\makeatother

\begin{document}
\title{Interpretable Intuitive Physics Model} 
\titlerunning{Interpretable Intuitive Physics Model}
\author{Tian Ye \inst{1} \and
Xiaolong Wang \inst{1} \and
James Davidson \inst{2} \and  Abhinav Gupta \inst{1}}

\authorrunning{Tian Ye, Xiaolong Wang, James Davidson, Abhinav Gupta}

\institute{Robotics Institute, Carnegie Mellon University \and Third Wave Automation}

\maketitle             
\begin{figure}
    \centering
    \includegraphics[trim={0 20pt 0 0},clip,width=0.8\linewidth]{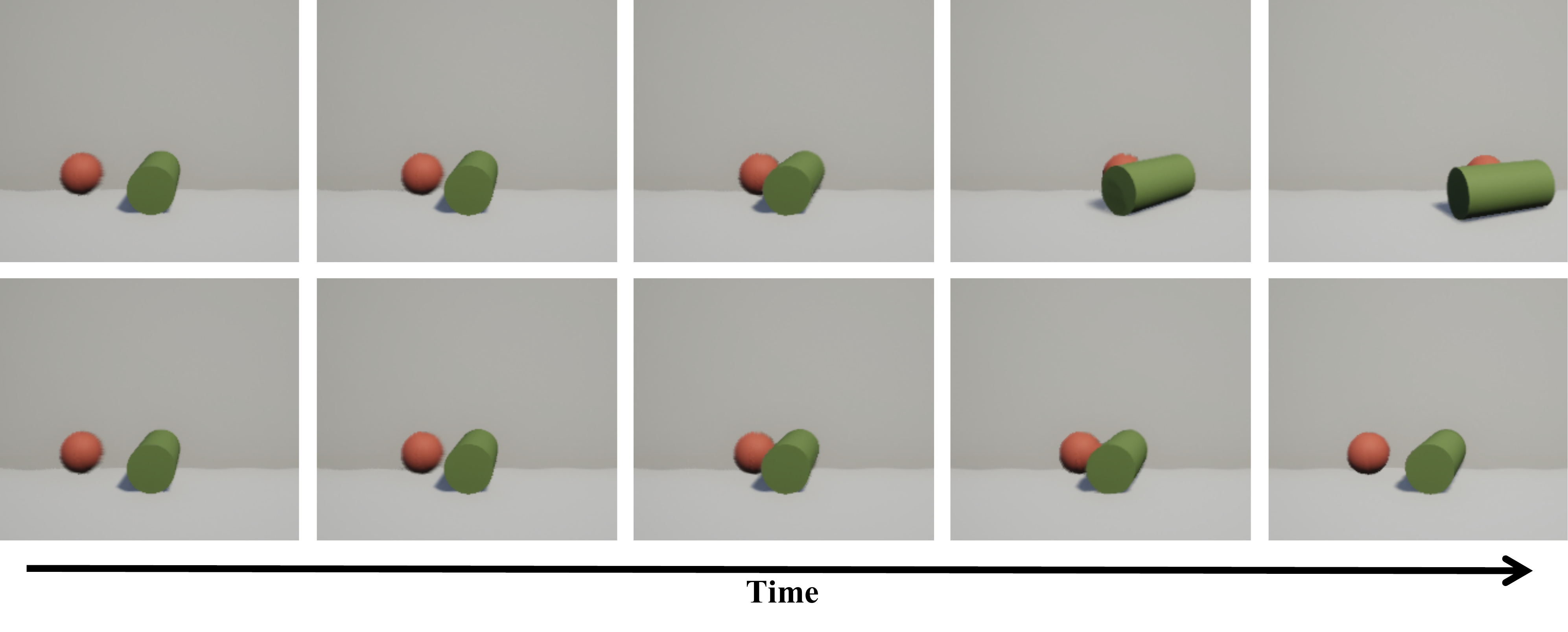}
    \caption{Interpretable Physics Models. Consider the sequences shown above. Not only we can predict the future frames of collisions but we can also predict the underlying factors that lead to such an inference. For example, we can infer the mass of cylinder is much higher in second sequence and therefore it hardly moves in the image. Our ability to infer meaningful underlying latent factors inspires us in this paper to learn an interpretable intuitive physics model.}
    \label{fig:teaser}
\end{figure}

\begin{abstract}
Humans have a remarkable ability to use physical commonsense and predict the effect of collisions. But do they understand the underlying factors? Can they predict if the underlying factors have changed? Interestingly, in most cases humans can predict the effects of similar collisions with different conditions such as changes in mass, friction, etc. It is postulated this is primarily because we learn to model physics with meaningful latent variables. This does not imply we can estimate the precise values of these meaningful variables (estimate exact values of mass or friction). Inspired by this observation, we propose an interpretable intuitive physics model where specific dimensions in the bottleneck layers correspond to different physical properties.  In order to demonstrate that our system models these underlying physical properties, we train our model on collisions of different shapes (cube, cone, cylinder, spheres etc.) and test on collisions of unseen combinations of shapes. Furthermore, we demonstrate our model generalizes well even when similar scenes are simulated with different underlying properties.
\keywords{Intuitive Physics \and Interpretable Models \and Physical Properties}
\end{abstract}

\section{Introduction}
Consider the collision image sequences shown in Figure~\ref{fig:teaser}. When people see these images, they not only recognize the shapes and color of objects but also predict what is going to happen. For example, in the first sequence people can predict that the cylinder is going to rotate while in the second sequence the ball will bounce with no motion on cylinder. But beyond visual prediction, we can even infer the underlying latent factors which can help us explain the difference in visual predictions. For example, a possible explanation of the behavior between the two sequences, if we knew the ball's mass didn't change, is that the first sequence's cylinder was lighter than the ball whereas in the second sequence the cylinder was heavier than the ball. Beyond this we can deduce that the cylinder in the first sequence was much lighter than the one in the second.

Humans demonstrate the profound ability to understand the underlying physics of the world~\cite{hamrick2011internal,hamrick2016inferring} and use it to predict the future. We use this physical commonsense for not only rich understanding but also for physical interactions. The question arises as to whether this physical commonsense is just an end-to-end model with intermediate representations being a black-box, or explicit and meaningful intermediate representations? For humans, the answer appears to be the latter. We can predict the future if some underlying conditions are changed. For example, we can predict that if we throw the ball in the second sequence with 10x initial speed then the cylinder might rotate.

In this paper, we focus on learning an intuitive model of physics~\cite{lerer2016learning,MottaghiECCV16,battaglia2016interaction}. Unlike some recent efforts, where the goal is to learn physics in an end-to-end manner with little-to-no constraints on intermediary layers, we focus on learning an {\bf interpretable} model. More specifically, the bottleneck layers in our network model physical properties such as mass, friction, etc.

Learning an interpretable intuitive physics model is, however, quite a challenging task. For example, Wu et al.~\cite{Wu17} attempts to build a model but the inverse graphics engine infers physical properties such as mass and friction. These properties are then used with neural physics engine or simulators for prediction. But can we really infer physical properties from the few frames of such collisions? Can we separate friction from mass, restitution by observing the frames? The fact is most of these physical factors are so dependent that it is infeasible to infer the exact values of physical properties. For example we can determine ratios between properties but not the precise values of both (e.g., we can determine the relative mass between two objects but not the exact values for both). This is precisely why in ~\cite{Wu17} only one factor is inferred from motion and the other factor is directly correlated to the appearance. Furthermore, the learned physics model is domain-specific and will not generalize--even across different shapes. 

To tackle these challenges, we propose an interpretable intuitive physics model, where specific dimensions in the bottleneck layers correspond to different physical properties. The bottleneck layer models the distribution rather than infer precise values of mass, speed and friction. In order to demonstrate that our system models these underlying physical properties, we train our model on collision of different shapes (cube, cone, cylinder, spheres etc.) and test on collisions of unseen combinations of shapes altogether. We also demonstrate the richness of our model by predicting the future states under different physical conditions (\eg, how the future frames will look if the friction is doubled). 

Our contributions include: (a) an intuitive physics model that disentangles different physical properties in an interpretable way; (b) 
a staggered training algorithm designed to distinguish the subtleties between different physical quantities; (c) generalization to different shapes and physical quantity combinations; most importantly, (d) the ability to adapt future predictions when physical environments change. Note (d) is different from generalization: the hallucination/prediction is done for a physical scene completely different from the observed first four frames.

\section{Related Work}

Physical reasoning and learning physical commonsense has raised a lot of interest in recent years~\cite{Zheng15,MottaghiECCV16,MottaghiCVPR16,Zhangcogsci16,Pinto16,Agrawal16,zhu2016inferring,edmonds2017feeling}. There has been multiple efforts to learn implicit and explicit models of physics commonsense. The underlying goal of most of these systems is to use physics to predict what is going to happen next~\cite{grzeszczuk1998neuroanimator,finn2016unsupervised,lerer2016learning,WenbinLi16,fragkiadaki2015learning,Wu15,Wu16}. The hope is that if the model can predict what is going to happen next after interacting with objects, it will be forced to understand the physical properties of the objects. For example,~\cite{lerer2016learning} tries to learn the physical properties by predicting whether a tower of blocks will fall. \cite{fragkiadaki2015learning} proposed to learn a visual predictive model for playing billiards. 

However, the first issue is what is the right data to learn this physics model. Researchers have tried a wide spectrum of approaches. For example, many researchers have focused on the task of visual prediction using real-world videos, based on the hypothesis that the predictive model will contain some underlying physical properties~\cite{vae_eccv2016,mathieu2015deep,vondrick2016generating}.  While videos provide realistic data, there is little to no control on how the data is collected and therefore the implicit models end up learning dynamic models of texture. In order to force physical commonsense learning, people have even tried using videos of physical interactions. For example, Physics101 dataset~\cite{Wu16} collects sequences of collisions for this task. But most of the learning still happens passively (random batches). In order to overcome that, recent approaches have tried to learn physics by active interaction using robots~\cite{Pinto16,Agrawal16,finn2016unsupervised}. While there is more control in the process of data collection, there are still issues with lack of diverse data due to most experiments being performed in lab setting with few objects. Finally, one can collect data with full control over several physical parameters using simulation. There has been lot of recent efforts in using simulation to learn physical models~\cite{lerer2016learning,fragkiadaki2015learning,MottaghiECCV16,MottaghiCVPR16}. One limitation of these approaches, in terms of data, is the lack of diversity during training, which forces them to learn physics models specific to particular shapes such as blocks, spheres etc. Furthermore, none of these approaches use the full power of simulation to generate a dense set of videos with multiple conditions. Most importantly, none of these approaches learn an interpretable model.

Apart from the question of data, another core issue is how explicit is the representation of physics in these models. To truly understand the object physical properties, it requires our model to be interpretable~\cite{Battaglia2016,Watters2017,Wu17,chang2016compositional,kulkarni2015deep}. That is, the model should not only be able to predict the futures, but the latent representations should also indicate the physical properties (e.g., mass, friction and speed) implicitly or explicitly. For example, ~\cite{Battaglia2016} proposed an Interaction Network which learns to predict the rigid body dynamics of gravitational systems. ~\cite{Wu17} proposed to explicitly estimate the physical object states and forward this state information to a physics engine for prediction. However, we argue exact values of these physical properties might not be possible due to entanglement of various factors. Instead of estimating the physics states explicitly, our work focuses on separating the dimensions in the bottleneck layer.

Our work is mostly related to the Inverse Graphics Network~\cite{kulkarni2015deep}. It learns a disentangled representation in the graphics code layer where different neurons  are encouraged to represent different transformations including pose and light. The system can be trained in an end-to-end manner without providing an explicit state value as supervisions for the graphics code layer. However, unlike the Inverse Graphics Network, where pose and light can be separately inferred from the input images, the dynamics are dependent on the joint set of physical properties in our model (mass, friction and speed), which confound future predictions. 

Our model is also related to the visual prediction models~\cite{kitani2012activity,vae_eccv2016,mathieu2015deep,Xue16,Zhoueccv16,qi2018future,Srivastava15} in computer vision. For example, \cite{Srivastava15} proposed to directly predict a sequence of video frames in raw pixels given a sequence of former frames as inputs. Instead of directly predicting the pixels,~\cite{vae_eccv2016} proposed to predict the optical flows given an input image and then warp the flows on the input images to generate future frames. However, the optical flow estimation is not always correct, introducing errors in the supervisions for training. To tackle this,~\cite{Zhoueccv16} proposed a bilinear sampling layer which makes the warping process differentiable. This enables them to train their prediction model from pixels to pixels in an end-to-end manner.  

\begin{figure}[!ht]
    \centering
    \includegraphics[width=0.95\linewidth]{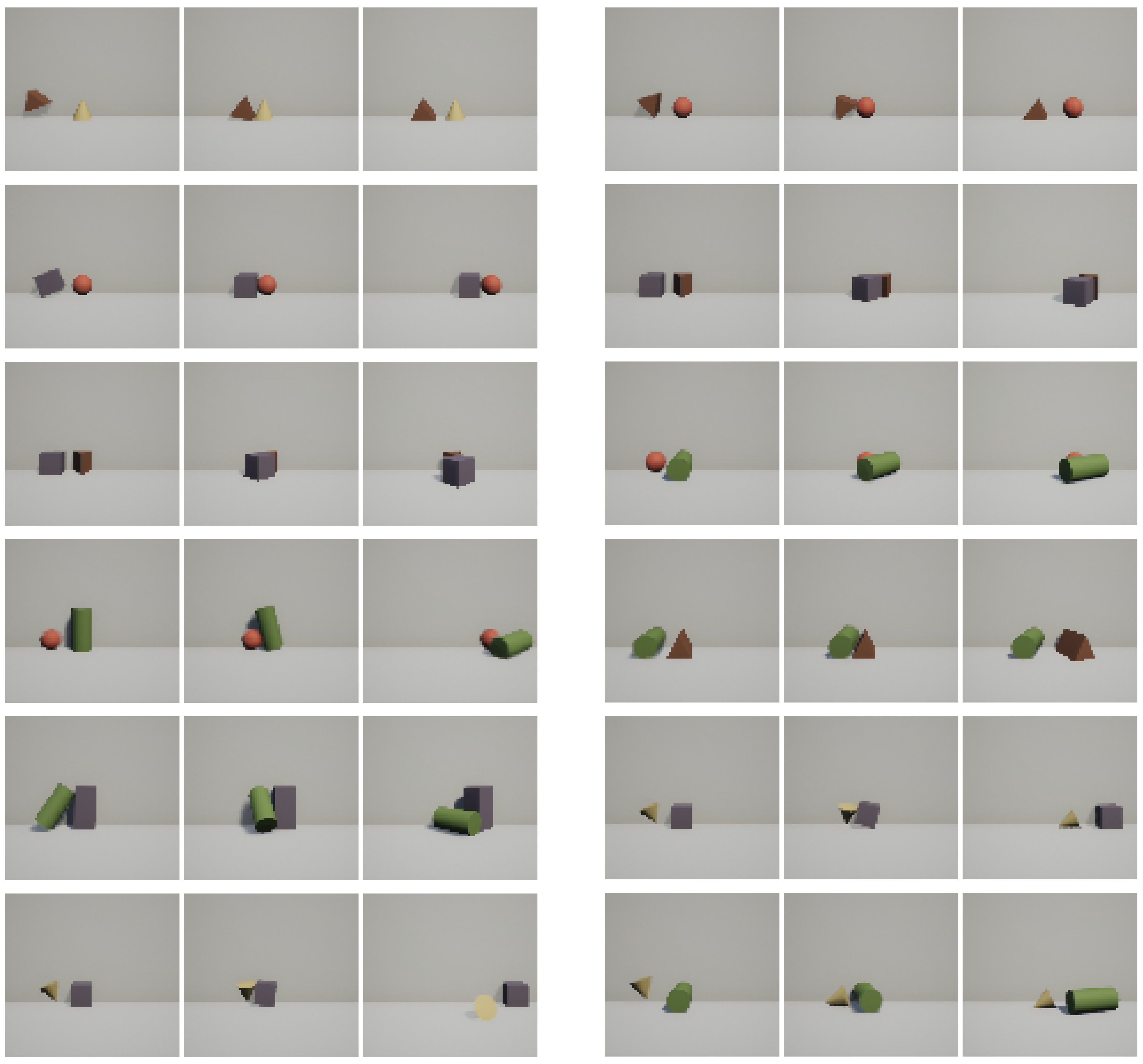}
    \label{fig:dataset}
    \caption{Our dataset includes 2 object collisions with a variety of shapes. Unlike existing physics datasets which have only one type of shape, our dataset is diverse in terms of different shapes and physical properties of objects.}
\end{figure}

\section{Dataset}
We create a new dataset for our experiments in this paper. The advantage of our proposed dataset is that we have rich combinations of different physical properties as well as different object appearances for different types of  collisions (falling over, twisting, bouncing, etc.). Unlike previous datasets, the physical properties in our dataset are independent from the object shapes and appearance. In this way, we can train models which force estimation of physical properties by observing the collisions. More importantly, our testing sets contain novel combinations of object shapes or physical properties that are unseen in the training set. The details of dataset generation is illustrated as following. 

We generate our data using the Unreal Engine 4 (UE4) game engine. We use 11 different object combinations with 5 unique basic objects: sphere, cube, cylinder, cone, and wedge. We select 3 different physical properties including mass of static object, initial speed of colliding object and friction of floor. For each property, we choose 5 different scales of values as shown in Table~\ref{tab:dataset}. For simplicity, we specify a certain scale of parameter by the format \{\emph{parameter name}\}$_{\{scale\}}$ (e.g., \emph{mass}$_1$, \emph{friction}$_4$, \emph{speed}$_2$). We simulate all the $5 \times 5 \times 5 = 125$ sets of physical combinations. For each set of physical property combination, there are 11 different object combinations and 15 different initial rotation and restitution. Thus in total there are $125 \times 15 \times 11=20625$ collisions. Each collision is represented by 5 sample frames with 0.5s time intervals between them. 

The diversity in our dataset is highlighted in Figure~\ref{fig:dataset}. For example, our dataset has cones toppling over; cylinders falling down when hit by a ball and rolling cylinders. We believe this large diversity makes it one of the most challenging datasets to learn and disentangle physical properties.

For training, we use $124$ sets of physics combination with $9$ different object combinations ($16740$ collisions). The remaining data are used for two types of testing: (i) parameter testing and (ii) shape testing. The parameter testing set contains $135$ collisions with unseen physical parameter combinations (\emph{mass}$_3$, \emph{speed}$_3$, \emph{friction}$_3$) but seen object shape combinations. The shape testing set on the other hand, contains $3750$ collisions with $2$ unseen shape combinations yet seen physical parameter combinations. We show the generalization ability of our physics model on both testing conditions. 

\begin{table}[t]
\centering
\small
\caption{Dataset Settings}
\tablestyle{6pt}{1.05}
\begin{tabular}{l|l|l|l|l|l}
\shline
~ & $\emph{scale}_{\ve{1}}$ & $\emph{scale}_{\ve{2}}$ & $\emph{scale}_{\ve{3}}$ & $\emph{scale}_{\ve{4}}$ & $\emph{scale}_{\ve{5}}$ \\
\shline
Mass & 100 & 200 & 300 & 400 & 500 \\
Speed & 10000 & 20000 & 30000 & 40000 & 50000\\
Friction & 0.01 & 0.02 & 0.03 & 0.04 & 0.05\\
\shline
\end{tabular}
\label{tab:dataset}
\end{table}

\section{Interpretable Physics Model}
Our goal is to develop a physics-based reasoning network to solve prediction tasks, \eg, physical collisions, while having interepretable intermediate representations. 

\subsection{Visual Prediction Model}
\begin{figure}[t]
    \centering
    \includegraphics[width=0.95\linewidth]{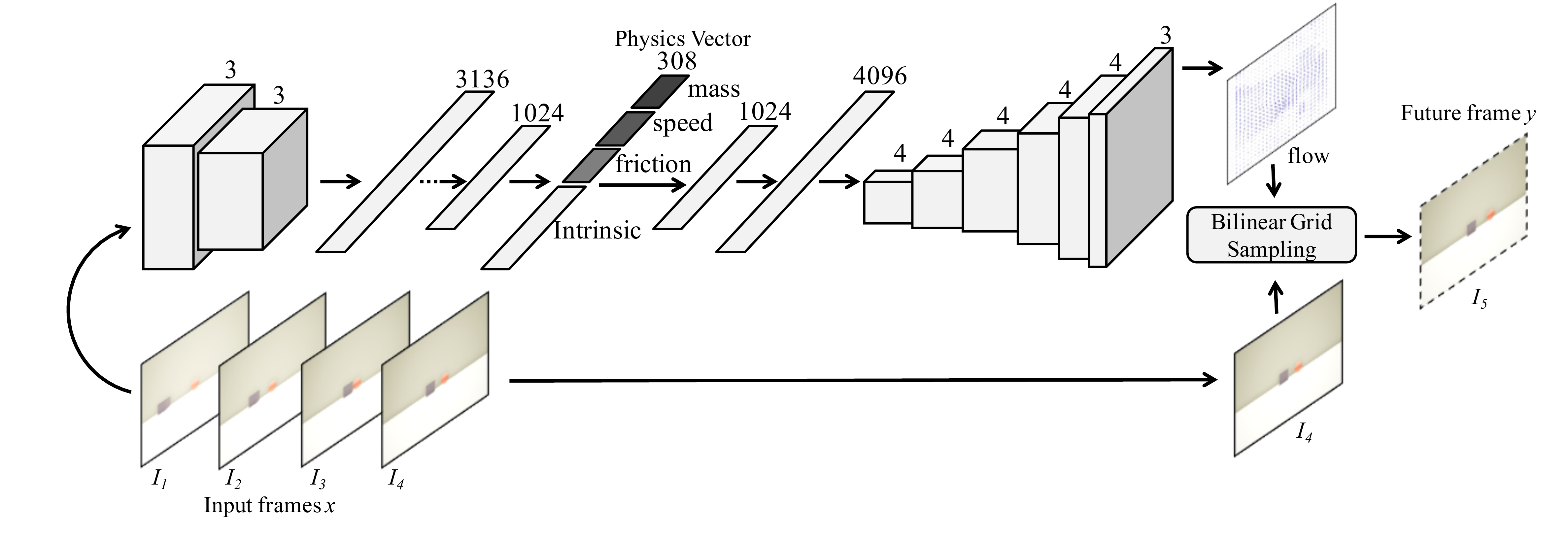}
    \caption{Model Architecture: we follow an encoder-decoder framework. The encoder takes 4 frames of a collision (2 before collision, 1 during collision, and 1 after collision). All inputs are first passed through a pre-trained Alexnet. The Alexnet features are further appended along channels and are sent to two convolution layers and four fully-connected layers. The resulting physics vector is passed through a decoder consisting of one fully-connected layer and six up-sampling convolution layers to produce an optical flow. The number on the convolution layers and transpose convolution layers stands for the kernel size of corresponding layer. The last bilinear grid sampling layer takes the optical flow and the $4^{th}$ input frame to produce future prediction.}
    \label{fig:model}
\end{figure}

As illustrated in Figure~\ref{fig:model}, our model takes in 4 RGB video frames as input and learns to predict the future 5th RGB frame after the collisions. The model is composed with two parts: an encoder for extracting abstract physical representations and a decoder for future frame prediction. 

\textbf{Encoder for physics representations.} The encoder is designed to capture the motion of two colliding objects, from which the physical properties can be inferred. Given 4 RGB frames as inputs, they are first forwarded to a ConvNet with AlexNet architecture and ImageNet pre-training. We extract the pool5 feature for each video frame and concatenate the features together as a representation for the input sequence. This feature is then forwarded to two convolutional layers and four fully connected layers to obtain the physics representation. 

The physics representation is a 306 dimensional vector, which contains disentangled neurons of mass (dimensions 1 to 25), speed (dimensions 26 to 50), friction (dimensions 51 to 75), and other intrinsic information (dimensions 76 to 306), as shown in Figure~\ref{fig:model}. Note that although the vector is disentangled, there is no explicit meanings for each neuron value. 

\textbf{Decoder for future prediction.} The physics representation is forwarded to a decoder for future frame prediction. Our decoder contains one fully-connected layer followed by six decovolutional layers. Inspired by~\cite{Zhoueccv16,vae_eccv2016}, our decoder uses optical flow fields as the output representation instead of directly outputing the RGB raw pixel values. The optical flow is then used to perform warping on the last input frame by a bilinear sampling layer~\cite{Zhoueccv16} to generate the future frame. Since the bilinear sampling layer is differentiable, the network can be trained in an end-to-end manner with the 5th frame for direct supervision. 

There are two major advantages of using optical flow as outputs: (i) it can force the model to learn the factors that cause the changes between two frames; (ii) it allows the model to focus on the changes of the foreground objects. 

\subsection{Learning Objective}

Formally, we define the encoder as a function $f$ and the decoder as a function $g$. Given an image sequence $x$ as inputs (4 frames), our encoder transforms the images into a physically meaningful and disentangled representation $z = f(x)$ and then the decoder transforms this representation into a future frame $y = g(z)$.

The disentangled representation $z$ can be formulated as $z = (\phi^{m}, \phi^{s}, \phi^{f}, \phi^{i})$ where  $(\cdot, \cdot)$ denotes concatenation. The first part $(\phi^{m}, \phi^{s}, \phi^{f})$ denotes the combination \textit{physics variable}, which encodes the physical quantities ($m$, $s$, $f$ stands for mass, speed, and friction respectively). The second part $\phi^{i}$ is the \textit{intrinsic variable}, representing all the other intrinsic properties in the scene (\eg, colors, shapes and initial rotation). 

In this paper, we study the effect of varying the values of physical quantities in a two-object collision scenario. Following the strategy  in~\cite{kulkarni2015deep}, we group our training sequence samples into mini-batches. Inside one mini-batch, only one physical property changes across all the samples and other physical properties remain fixed. We denote $B^{p}=\{(x_k, y_k)\}_{k=1}^{5}$ as one mini-batch with 5 sequences, where the only changing property is $p$ (i.e., we use $p$ as a variable to represent either mass, speed or friction). 

For each mini-batch $B^p$ during training, we encourage only the dimensions corresponding to the property $p$ to change in $z$. For example, when training with a mini-batch where only mass is changing, we force the network to have different values in the dimensions for $\phi^{m}$ and same values for the rest of the dimensions in $z$. For simplicity, we further denote the dimensions which relevant to $p$ in $z$ as $\phi^{p}_k$ and the rest of the dimensions as $\bar{\phi^{p}_k}$ for example $k$.

We train our prediction model with this constraint. Assuming we are training with one batch $B^{p}=\{(x_k, y_k)\}_{k=1}^{5}$. In a maximum likelihood estimation (MLE) framework, this can be formulated as maximizing the log-probabilities under the desired constraints:
\begin{equation}
\begin{aligned}
& {\text{maximize}}
& & \sum_{k=1}^{5} \log(\ma{P}(y_k | x_k)) \\
& \text{subject to}
& & \bar{\phi^{p}_i} = \bar{\phi^{p}_j}, \forall 1 \leq i,j \leq 5
\end{aligned}
\label{eqn:mle}
\end{equation}
where $\bar{\phi^{p}_k}$ contains both the intrinsic variable inferred from image sequence $x_k$ and inferred physics variables, except for the changing parameter. 

In our auto-encoder architecture, the objective function is equivalent to minimizing the l1 distance between the predicted images $\hat{y}_{k}$ and the ground truth future images $y_{k}$: 
\begin{equation}
\begin{aligned}
\mathcal{L}_{mle} = \sum_{k} ||\hat{y}_{k} - y_{k}||_1. 
\end{aligned}
\label{eqn:loss_mle}
\end{equation}

The constraints in Eq.~\ref{eqn:mle} can be satisfied via minimizing the loss between $\bar{\phi^{p}_k}$ and the mean of them within the mini-batch $\bar{\phi^{p}} = \frac{1}{5} \sum_k \bar{\phi^{p}_k}$ as, 
\begin{equation}
\begin{aligned}
\mathcal{L}_{ave} =\sum_{k} ||\bar{\phi^{p}_k} - \bar{\phi^{p}}||_2^2.
\end{aligned}
\label{eqn:loss_ave}
\end{equation}

We apply both losses jointly during training our model with a constant $\lambda$ balancing between them as,
\begin{equation}
\begin{aligned}
\mathcal{L} = {L}_{mle} + \lambda {L}_{ave}. 
\end{aligned}
\label{eqn:loss_sum}
\end{equation}
In practice, we set the $\lambda$ dynamically so that both gradients are maintained in the same magnitude. The value of $\lambda$ is around $1e-6$.   

\subsection{Staggered Training}

Although we follow the training objective proposed in~\cite{kulkarni2015deep}, it is actually non-trivial to directly optimize with this objective. There is a fundamental difference between our problem and the settings in~\cite{kulkarni2015deep}: the physical dynamics are dependent across the set of properties, which confounds training. The same sequence of inputs and output ground-truth might infer different combinations of the physical properties. For example, both large friction and slow speed can lead to small movements of the second object after collision. Thus modifications on training method is required to handle this multi-modality issue. 

We propose a staggered training algorithm to alleviate this problem. We first divide the entire training set $D$ into 3 different sets $\{D^{p}\}$, where $p$ indicates one of the physics properties( mass, speed or friction). Each $D^p$ contains different mini-batches of $B^p$, inside which the only changing property is indicated by $p$. 

The idea is: instead of training with all the physics properties at the same time in the beginning, we perform curriculum learning. We first train the network with one subset $D^p$ and then progressively add more subsets with different properties into training. In this way, our training set becomes larger and larger through time. By learning the physics properties in this sequential manner, we force the network to recognize new physical properties one by one while keeping the learned properties. In practice, we observe that in the first training session, the network behaves normally. For the following training sessions, the loss will increase in the beginning, and will decrease to roughly the same level as the previous session. 

\section{Experiments}

We now demonstrate the effectiveness and generalization of our model. We will perform two sets of experiments with respect to two different testing sets in our dataset. One tests on unseen physical property combinations but seen shape combinations, and the other tests on unseen shape combinations with seen physical properties. Before going into further analysis, we will first describe the implementation details of our model and the baseline method. 

\noindent\textbf{Implementation details} In total, we trained for 319 epochs. We used ADAM for optimization, with initial learning rate $10^{-6}$. During training, each mini-batch mentioned above has 5  sequences. During the training for the first physical quantity, each batch contains 3 mini-batches, which means 15 data in total. For the second round of staggered training, each batch contains 2 mini-batches, one for each physical quantity; similarly, in the third round of training, each batch contains 3 mini-batches, one for each physical quantity.

\noindent\textbf{Baseline model} Our baseline model learns intuitive physics in an end-to-end manner and post-hoc obtains the dimensions that correspond to different physical properties. We need the disentagled representation because we want to test the generalization when the physical properties are different from input video: \eg, what happens if friction is doubled? What happens if the speed is 1/10th?

For the baseline, we use the same network architecture. Different from our approach, we do not add any constraints on the bottleneck representation layer as in Eq.~\ref{eqn:mle} in the baseline model. However, we still want to obtain the disentangled representation from this baseline for comparison. Recall that we have a subset $D^p$ for each property $p$ (mass, friction or speed). The examples in each mini-batch inside $D^p$ specify the change of property $p$. We compute the variances for each neuron in the bottleneck representation for each $D^p$, and select 25 dimensions with top variances as the vector indicating property $p$. 
\begin{figure}[t]
    \centering
    \includegraphics[width=0.85\linewidth]{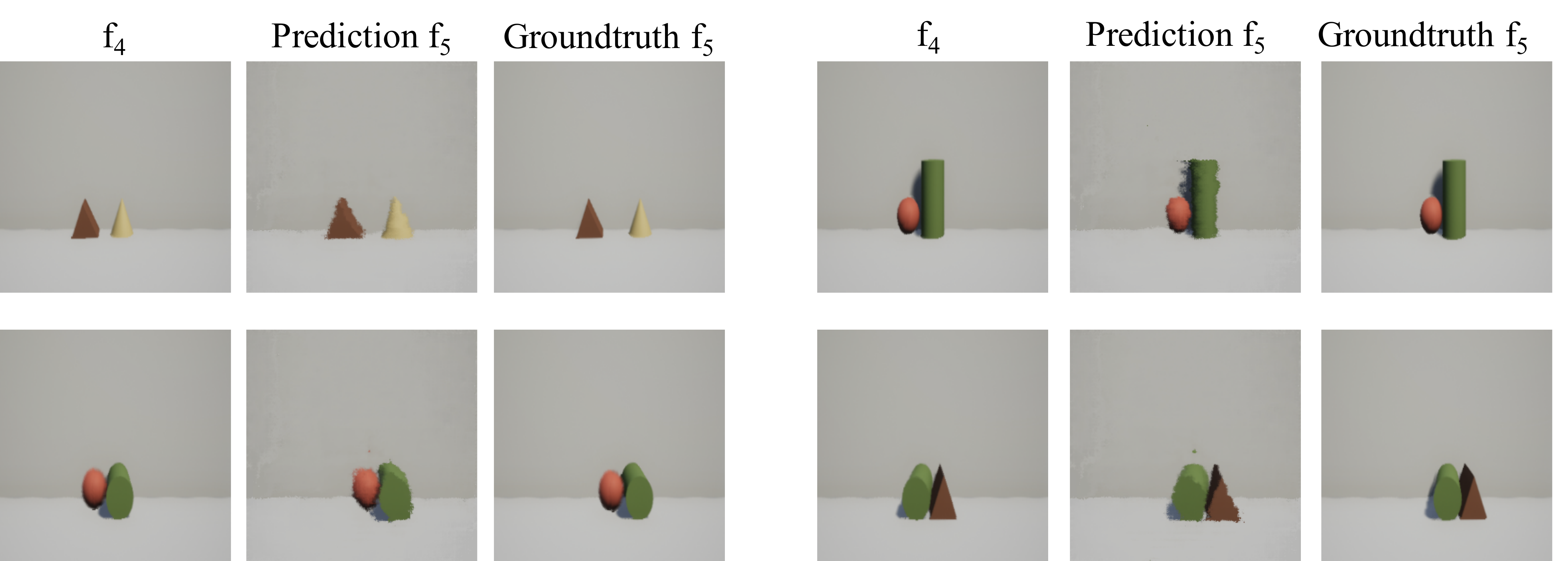}
    \caption{Prediction results for unseen parameters but seen shapes.}
    \label{fig:output_parameter}
\end{figure}

\begin{figure}[t]
    \centering
    \includegraphics[width=0.85\linewidth]{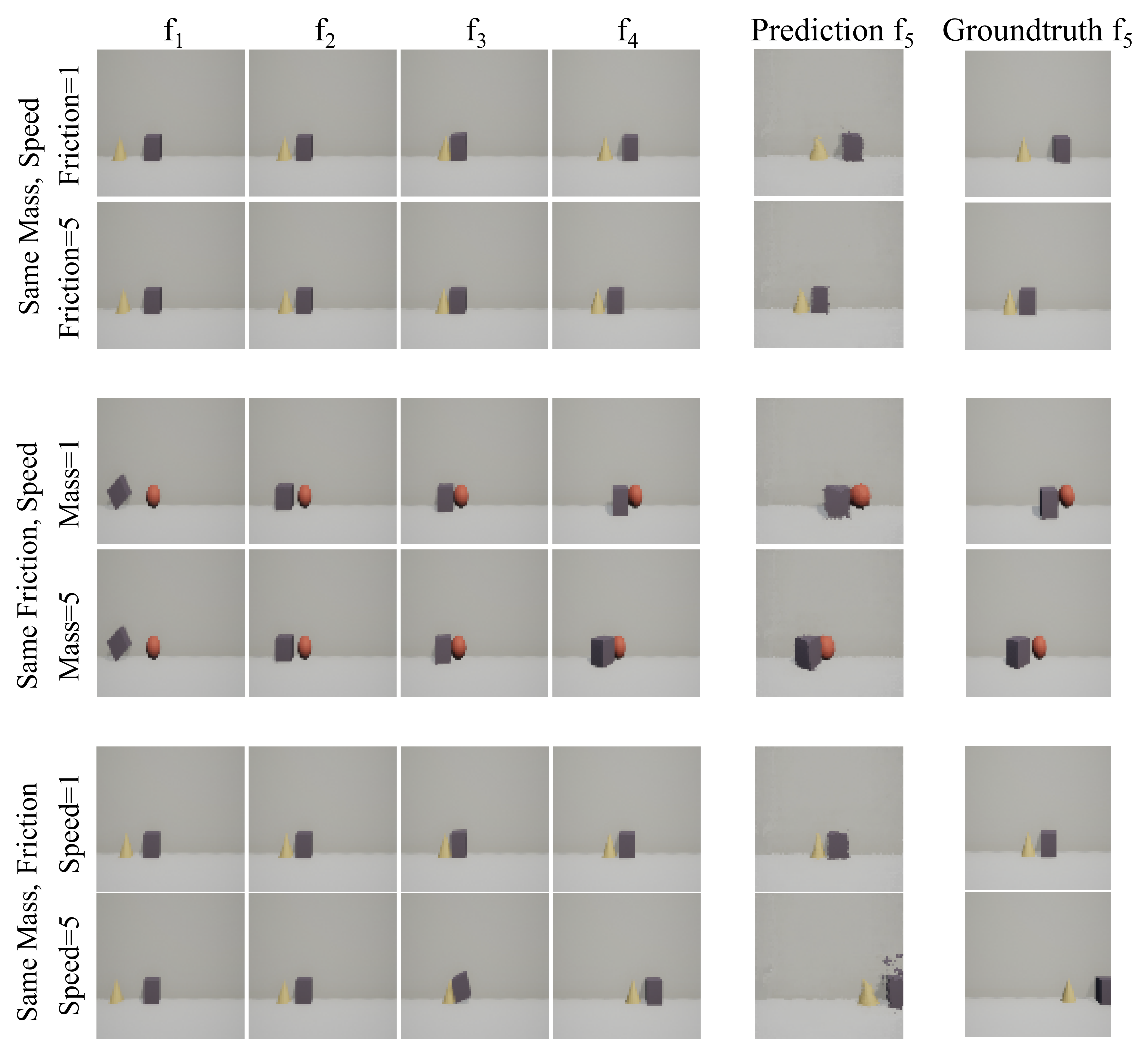}
    \caption{4 input frames, the predicted 5th frame and ground-truth for collisions with unseen shape combinations. Contrast the predictions as one of physical property changes. For example, to show our approach understand these shapes, we predict for two different friction values in first case (keeping mass and speed same). The less motion in 2nd case shows that our approach understands the concept of friction.}
    \label{fig:output_sequences}
\end{figure}

\begin{figure}[t]
    \centering
    \includegraphics[width=0.9\linewidth]{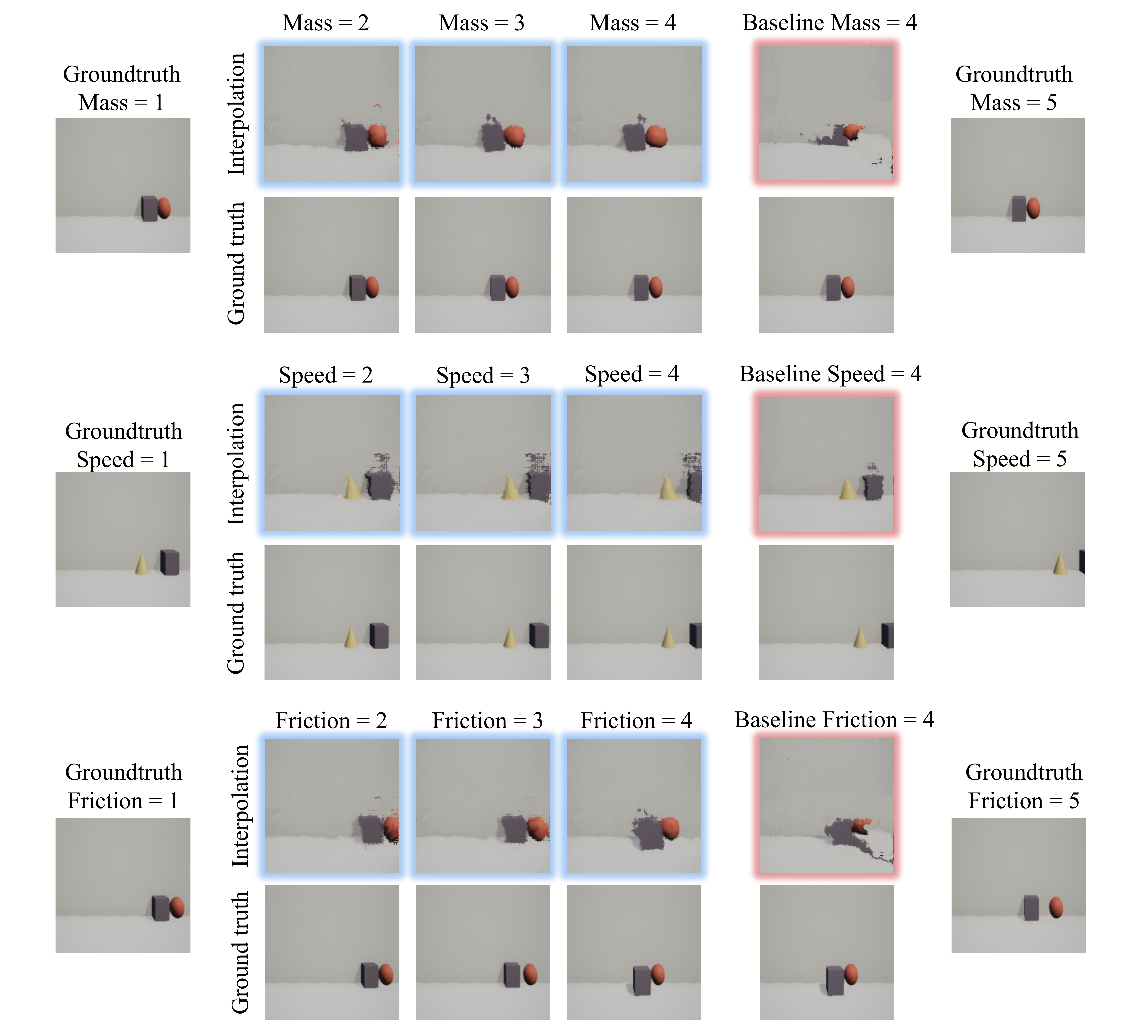}
    \caption{Interpolation results for physical quantity with different values. Our interpolation results are shown with blue frames. Images with red frame in last column represents the interpolation results for baseline when physical quantities equal to 4.}
    \label{fig:interpolate_result}
\end{figure}

\subsection{Visual prediction}
\noindent {\bf Unseen Parameters:} First we evaluate if we can predict future pixels when we see a novel combination of physical parameters. Specifically, our model has never seen in training a combination of mass=3, friction=3 and speed=3. Figure~\ref{fig:output_parameter} shows our interpretable model generalizes well and produces high quality predictions.

\noindent {\bf Unseen Shape Combinations:} Next, we want to explore if our visual prediction model generalizes to different shape combinations using two unseen sets: (a) cone and cuboid; (b) cuboid and sphere. To demonstrate that our model understands each of these physical properties, we show contrasted prediction results for two different values. For example, we will use different friction values ($1,5$) but same mass and speed. Comparing these two outputs should highlight how our approach understands the underlying friction values.

As shown in Figure.~\ref{fig:output_sequences}, our predicted future frame has high quality compared to the ground-truth. We show that our model can generalize the physics reasoning to unseen objects and learn to output different collisions results given different physical environments. For example in the second condition, when the mass of sphere is high ($5$), our approach can predict it will not move and instead the cube will bounce back. We also compare our approach to baseline quantitatively: our approach has pixel error of 87.3, while baseline has pixel error of 95.6.The results clearly indicate our interpretable model tends to generalize better than an end-to-end model when test conditions are very different. 

In addition to the baseline, we also compare our model with two other methods based on optical flow. First, we trained another prediction network using the optical flow computed between the 4th and the 5th frame as direct supervisions, instead of using the pixels of the 5th frame. For testing, we apply the predicted optical flows on the 4th frame to generate the future frame. The loss between the future frame and the ground-truth 5th frame is 118.8. Second, we computed 3 optical flows of first 4 frames, using which to find a linear model to generate the future optical flow. We apply this optical flow on the 4th frame and compare the result to the ground-truth 5th frame. The error reaches to 292.5. The result shows that our method achieves high precision than using optical flow directly.

\subsection{Physical Interpolation}

To show our model has actually learnt physics properties, we perform a series of interpolations on the bottleneck representation. 

\noindent\textbf{Interpolating physics representation within a mini-batch.} We first show that the learned bottleneck layer is meaningful and smooth. To demonstrate this, we interpolate between different physical properties and compare our result with the ground-truth. The experiment is conducted in the following way. Let's take mass as an example: given a mini-batch where only mass changes, we use the encoder to get the physics vector $z_1 = (\phi^m_1, \phi^s_1, \phi^f_1, \phi^i_1)$ from mass$_1$ data and $z_5 = (\phi^m_5, \phi^s_5, \phi^f_5, \phi^i_5)$ from mass$_5$ data. To estimate the physics vector for mass$_i$, we interpolate a new mass variable $\hat{\phi}^m_i = (1-0.25i) \cdot \phi^m_1 + 0.25i \cdot \phi^m_5$ and use this to create a new physics vector $\hat{z}_i = (\hat{\phi}^m_i, \phi^s_1, \phi^f_1, \phi^i_1)$. We pass the new vector to the decoder to predict the optical flows, which are warped to the 4th image in sequence $i$ via the bilinear sampling layer, and generate the future frame. 

We perform the same set of experiments for the baseline model. Quantitatively, we evaluate the prediction using the sum of mean square error for each pixel, as shown in Table~\ref{tab:interpolate_result}, which shows that our method is significantly better than the baseline. We also visualized the results in Figure~\ref{fig:interpolate_result}. Interestingly, our interpolation results are also very close to the ground-truth. On the other hand, baseline models failed easily when there is a dramatic change during interpolations.

We also trained another model which takes physics parameters and the optical flows of first 4-frame as inputs, and predicts the future frame. This model performs much worse than our model in the interpolation test as shown in Figure~\ref{fig:interpolate_result}. 
We believe a ground-truth physics parameter based approach focuses on classification instead of learning an intuitive physics model. 
In interpolation experiments, the model cannot separate physics information from the optical flow features. 

From these comparison, we can see that only by learning interpretable representations, we can generate reasonable prediction results after interpolations.  

\begin{table}[!t]
\caption{Interpolation Result. The numbers are pixel prediction errors}
\begin{center}
\resizebox{0.75\textwidth}{!}{
\begin{tabular}{l||c|c|c|c||c}
\Xhline{2\arrayrulewidth}
Method & shape 2 & shape 3 & shape 4 & shape 5 & parameter 3\\
\hline
Baseline & 117.76 & 130.41 & 154.78 & 173.80 & 299.88\\
\hline
Flow + Physics & 272.02 & 317.79 & 328.06 & 336.54 & 671.51 \\
\hline
Ours & \textbf{110.93} & \textbf{119.73} & \textbf{131.70} & \textbf{138.04} & \textbf{154.09}\\
\Xhline{2\arrayrulewidth}
\end{tabular}
}
\label{tab:interpolate_result}
\end{center}
\end{table}

\begin{figure}[!t]
    \centering
    \includegraphics[width=0.7\linewidth]{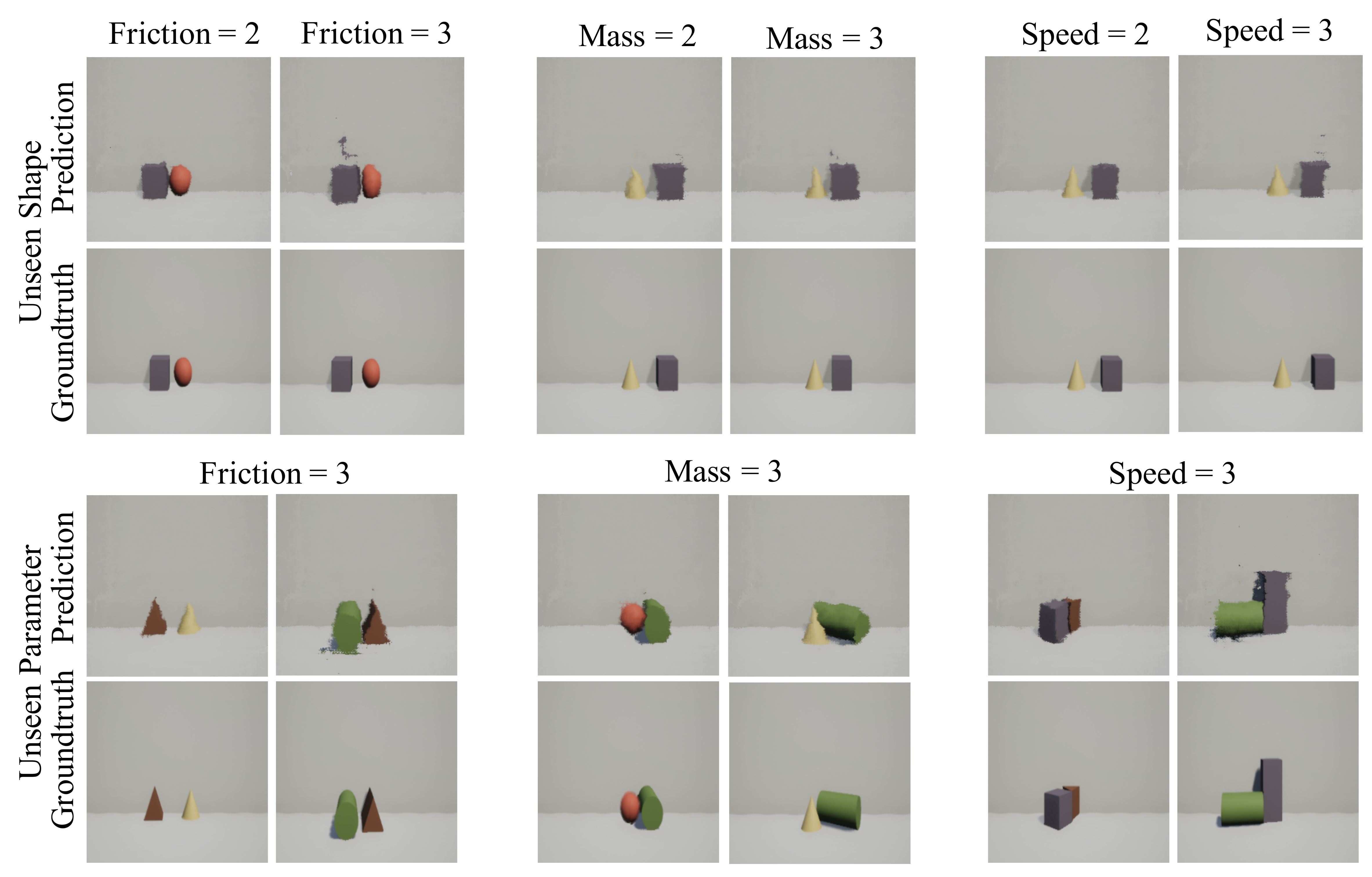}
    \caption{Prediction by learning double, triple ratio relation for different physical entities. Top: the result with unseen shapes. Bottom: result with unseen parameters.}
    \label{fig:ratio}
\end{figure}

\noindent\textbf{Changing physical properties.} In this experiment, we show that physics variables learned by our model are interpretable by finding a mapping between different scale of the same physical property. Specifically, we want to see: can we predict the future if the mass is doubled while all other physics conditions remain the same? For each physical quantity $p$, we train two networks $F^p_2$ and $F^p_3$ which learns to double or triple the scale of a physical property. For example, we can project the physics representation of mass$_1$ to mass$_3$ by using the network $F^p_3$. The network architecture for both $F^p_2$ and $F^p_3$ is a simple 2-layer fully connected network with 256 hidden neurons per layer. These two networks can be trained using the physical representations inferred by our encoder with the training data. 

In testing time, we apply the similar interpolation as the last experiment. The only difference is that instead of using an interpolation between two relevant representations, we use the fully connected network to generate the new representations. We again evaluate the quantitative results by computing the mean square error over the pixels. As shown in Table~\ref{tab:ratio_result}, we have a larger performance gain in this setting compared to the baseline. Figure~\ref{fig:ratio} shows the prediction results of our model when the physics property is enlarged from scale $1$ to $2$ and $3$, which are all very close to the ground-truth. This is another evidence showing our physics representation is interpretable and generalizes significantly better.

\begin{table}[!t]
\caption{Ratio Result. Comparing visual prediction when underlying physical parameters are changed by a factor}
\begin{center}
\resizebox{0.75\textwidth}{!}{
\begin{tabular}{l||c|c||c}
\Xhline{2\arrayrulewidth}
Method & shape ratio 2 ($\downarrow$) & shape ratio 3 ($\downarrow$) & parameter ratio 3 ($\downarrow$) \\
\hline
\hline
Baseline & 345.60 & 310.37 & 490.92 \\
\hline
Ours & \textbf{110.79} & \textbf{124.00} & \textbf{157.10}\\
\Xhline{2\arrayrulewidth}
\end{tabular}
}
\label{tab:ratio_result}
\end{center}
\end{table}

\begin{figure}[!t]
    \centering
    \includegraphics[width=0.7\linewidth]{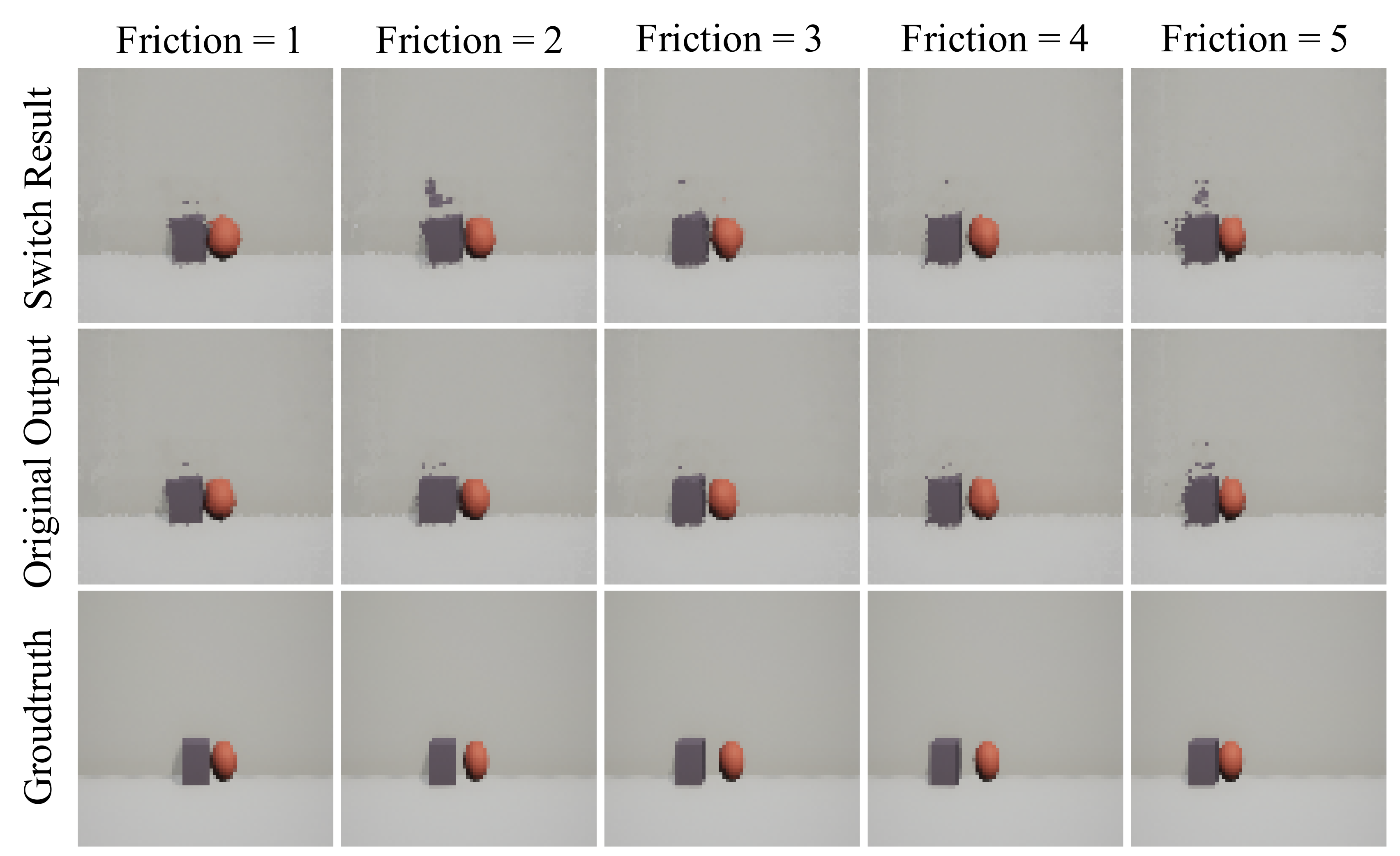}
    \caption{Prediction when physical property vector from one shape combination is applied to a different shape combinations. The first row shows switched result; the second row shows the prediction without switching; the third row shows ground-truth.}
    \label{fig:switch_friction}
\end{figure}

\noindent\textbf{Switching between the object shapes.} In experiments above, we interpolate the physics representation and apply them to the same object shape combinations. In this experiment, for a physical property $p$, we replace the corresponding variable $\phi^p$ of one collision with the variable from another collision with different objects but the same $p$ value. We visualize the results in Figure~\ref{fig:switch_friction}, where the first line shows the predictions when we replace current $\phi^p$ with one from another shape combination. The results are almost same as the original prediction and the ground-truth, which means that the physical variable of same value can be transferred among different shape combinations. It also shows that the dimensions of physics and other dimensions are independent and can be appended easily.

\section{Conclusions}
We demonstrated an interpretable intuitive physics model that generalizes across scenes with different underlying properties and object shapes. Most importantly, our model is able to predict the future when physical environment changes. To achieve this we proposed a model where specific dimensions in the bottleneck layers correspond to different physical properties. However, often physical properties are dependent and intertangled, so we introduced a training curriculum and generalized loss function that was shown to outperform the baseline approaches.

{\noindent {\bf Acknowledgement}: Research was sponsored by the Army Research Office and was accomplished under Grant Number W911NF-18-1-0019. The views and conclusions contained in this document are those of the authors and should not be interpreted as representing the official policies, either expressed or implied, of the Army Research Office or the U.S. Government. The U.S. Government is authorized to reproduce and distribute reprints for Government purposes notwithstanding any copyright notation herein. We would like to thank Yin Li and Siyuan Qi for helpful discussions.}

\bibliographystyle{splncs04}
\bibliography{1443}
\end{document}